\documentclass[sigconf,nonacm]{acmart}
\usepackage{url}
\usepackage{bm}
\usepackage{booktabs}
\usepackage{multirow}
\hypersetup{colorlinks=true, linkcolor=blue, anchorcolor=red, citecolor=blue, filecolor =red, pdfauthor=author}
\usepackage{xcolor}
\usepackage{color, colortbl}
\newcommand{\bb}{\color{black}}

\definecolor{rosso}{RGB}{249,128,115}

\usepackage[table]{xcolor}

\definecolor{pv}{HTML}{4F6272}       
\definecolor{weath}{HTML}{C25E5E}     
\definecolor{fused}{HTML}{E3B04B}  
\definecolor{forecast}{HTML}{5C8D4C}
\definecolor{lightblue}{HTML}{E6E5FD}
\definecolor{lightgreen}{RGB}{196, 227, 190}
\definecolor{lightred}{RGB}{239, 172, 170}
\definecolor{lightpurple}{RGB}{186, 212, 229}
\definecolor{lightyellow}{RGB}{244, 202, 141}
\definecolor{violet}{RGB}{128,0,128}

\usepackage{xcolor}
\usepackage{tikz}

\definecolor{gold}{HTML}{FBF2D2}
\definecolor{silver}{HTML}{DDDDDD}
\definecolor{bronze}{HTML}{EED2B8}

\definecolor{goldD}{HTML}{D9AE13}
\definecolor{silverD}{HTML}{909090}
\definecolor{bronzeD}{HTML}{9A5F26}

\newcommand{\medal}[3]{%
  \tikz[baseline=(char.base)]{%
    \node[
      rounded corners=2pt,
      fill=#1,
      draw=#2,
      inner sep=1.5pt
    ](char){#3};%
  }%
}

\newcommand{\rk}[2]{%
  \ifcase#1\or
    \medal{gold}{goldD}{\textbf{#2}}%
  \or
    \medal{silver}{silverD}{#2}%
  \or
    \medal{bronze}{bronzeD}{#2}%
  \else
    #2%
  \fi\ignorespaces
}

\definecolor{lightblue}{HTML}{E6E5FD}

\usepackage{utfsym}

\usepackage{pifont} 

\AtBeginDocument{%
  }

\setcopyright{acmlicensed}
\copyrightyear{2026}
\acmYear{2026}
\setcopyright{cc}
\setcctype{by-nc-nd}
\acmConference[SIGSPATIAL '26]{The 34th ACM International Conference on Advances in Geographic Information Systems}{November 03--06, 2026}{Riverside, CA, USA}
\acmBooktitle{The 34th ACM International Conference on Advances in Geographic Information Systems (SIGSPATIAL '26), November 03--06, 2026, Riverside, CA, USA}
\acmDOI{10.1145/3841645.3843356}
\acmISBN{979-8-4007-2950-8/2026/11}

\usepackage{lipsum}

\begin{document}

\title{VegSim: A Geospatial World Model for Scenario-Conditioned Vegetation Simulation
}

\author{Irene Iele}
\orcid{0009-0001-1344-0609}
\affiliation{%
  \institution{Università Campus Bio-Medico di Roma}
  \city{Rome}
  \country{Italy}
}
\email{irene.iele@unicampus.it}

\author{Elena {Mulero Ayllón}}
\orcid{0009-0005-8969-7901}
\affiliation{%
  \institution{Università Campus Bio-Medico di Roma}
  \city{Rome}
  \country{Italy}
}
\email{e.muleroayllon@unicampus.it}

\author{Paolo Soda}
\orcid{0000-0003-2621-072X}
\affiliation{%
  \institution{Umeå University}
  \city{Umeå}
  \country{Sweden}
}
\additionalaffiliation{%
  \institution{Università Campus Bio-Medico di Roma}
  \city{Rome}
  \country{Italy}
}
\email{paolo.soda@umu.se}


\author{Matteo Tortora}
\correspondingauthor
\orcid{0000-0002-3932-7380}
\affiliation{%
  \institution{ University of Genoa}
  \city{Genoa}
  \country{Italy}
}
\email{matteo.tortora@unige.it}

\renewcommand{\shortauthors}{Iele et al.}

\begin{abstract}
Vegetation monitoring under climate stress requires answering not only how it will evolve given the expected weather, but how it would respond to alternative meteorological conditions.
Forecasting models return the expected vegetation state for the observed weather and cannot answer these scenario-conditioned questions, because future weather is fixed to the recorded trajectory.
We present VegSim, a geospatial world model for scenario-conditioned vegetation simulation.
VegSim infers a latent vegetation state from sparse satellite-derived NDVI histories, past meteorological covariates, and static spatial context, propagates it forward under future weather forcing through recurrent latent dynamics, and decodes predictive NDVI quantiles at each lead time.
Because future forcing enters as a controllable input, the same trained model supports probabilistic forecasting under observed weather and conditional simulation under user-defined meteorological forcing, without supervision on scenario responses.
We evaluate VegSim on GreenEarthNet across in-distribution data and spatial, temporal, and joint spatio-temporal shifts, where it achieves strong point and probabilistic accuracy against time series and Earth Observation forecasting baselines while using a compact architecture.
We then simulate vegetation responses across Europe under two contrasting meteorological scenarios, obtaining a weak but broadly positive response under winter wet warming and a widespread negative response under summer drought.
The code is available at~\url{https://github.com/arco-group/vegsim}.
\end{abstract}




\maketitle

\begin{figure*}
    \centering
\includegraphics[width=\linewidth]{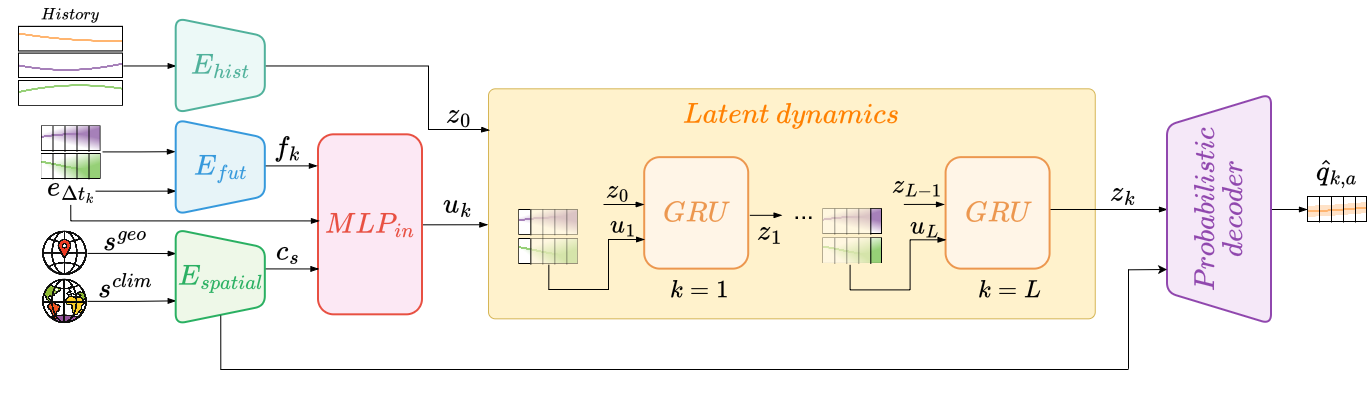}
   \caption{Overview of VegSim. The history encoder maps sparse historical NDVI and meteorological observations to an initial latent vegetation state $\mathbf{z}_0$. Future meteorological forcing, lead-time information, and static spatial context define the per-step inputs $\mathbf{u}_k$ that drive the recurrent latent dynamics. A shared probabilistic decoder maps each latent state $\mathbf{z}_k$ to predictive NDVI quantiles $\hat{\mathbf{q}}_{k,a}$.}
\label{fig:architecture}
    \Description{Schematic of the VegSim architecture. Sparse historical NDVI observations and meteorological covariates are processed by the history encoder to infer an initial latent vegetation state. Dense future meteorological forcing is combined with lead-time embeddings and processed by the future encoder to obtain per-step future representations. Geographic coordinates and the K\"oppen--Geiger climate class are encoded into a static spatial context vector. At each rollout step, the future representation, spatial context, and lead-time embedding are combined to form the input to recurrent GRU dynamics, which propagate the latent vegetation state from the initial state across the forecast horizon. A shared probabilistic decoder combines each latent state with the spatial context to produce predictive NDVI quantiles.}
    \label{fig:architecture}
\end{figure*}

\section{Introduction}
Vegetation monitoring under weather and climate stress requires answering not only how vegetation will evolve under expected conditions, but also how it would respond to alternative meteorological trajectories. Such questions arise in drought assessment, heatwave analysis, and climate-risk monitoring, where the effect of warmer, drier, or colder conditions may be as relevant as the most likely future trajectory.
In agricultural applications, predictive models of environmental dynamics can also serve as simulation components within emerging agentic decision-support systems~\cite{bonifacio2026agentic}.
Earth Observation (EO) provides repeated measurements of vegetation state, while meteorological products provide the external drivers that shape its evolution. Modeling their interaction, however, is complicated by the sparse and irregular availability of clear-sky satellite observations and by the dependence of vegetation response on season and location.
Recent forecasting models increasingly combine historical observations with exogenous meteorological covariates across environmental applications \cite{tortora2026matnet}.
Contextformer combines satellite observations, meteorological covariates, and static spatial information to forecast vegetation under spatial and temporal distribution shifts~\cite{benson2024multimodal}.
Probabilistic approaches further model uncertainty in future vegetation trajectories~\cite{iele2026probabilistic,zhao2025vegediff}. 
Other studies perturb weather inputs to analyze predictor sensitivity or project vegetation under climate scenarios~\cite{diaconu2022understanding,chen2022deep}. 
These approaches nevertheless remain forecasting models: future meteorological inputs are used to predict the corresponding observed trajectory, while perturbations are typically introduced as an auxiliary analysis.
In parallel, world models learn latent dynamical systems whose internal state is propagated under external conditioning signals~\cite{ha2018recurrent,hafner2025mastering}.
Recent work has begun to transfer this paradigm to remote sensing~\cite{xu2026rs}, but controllable meteorological forcing for vegetation simulation remains largely unexplored.

We introduce VegSim, a geospatial world model for scenario-conditioned vegetation simulation. 
VegSim infers a latent vegetation state from sparse historical NDVI observations, past meteorological covariates, and static spatial context, then propagates this state through recurrent latent dynamics conditioned on future weather forcing.
A probabilistic decoder produces NDVI quantiles at each lead time.
Because the future forcing is an explicit model input, the same trained model supports \emph{forecasting} under observed forcing and \emph{scenario simulation} under user-defined perturbations by replacing the future forcing sequence.
Our contributions are:
\begin{itemize}
\item we formulate vegetation simulation as probabilistic latent rollout under controllable meteorological forcing;
\item we introduce VegSim, a compact geospatial world model that combines sparse EO histories, dense weather drivers, and static spatial context;
\item we evaluate its forecasting performance across in-distribution and spatial, temporal, and joint spatio-temporal shifts, and demonstrate scenario-conditioned vegetation responses under controlled meteorological perturbations across Europe.
\end{itemize}

\section{VegSim}
\label{sec:vegsim}

\paragraph{\textbf{Data and representation.}}
We evaluate VegSim on GreenEarthNet~\cite{benson2024multimodal}, a continental-scale EO dataset organized into spatio-temporal minicubes over Europe.
Each minicube contains 30 Sentinel-2 observations sampled every five days over $128\times128$ pixels at 20\,m resolution, together with 150 daily meteorological observations.
The meteorological drivers include wind speed, relative humidity, shortwave downwelling radiation, rainfall, sea-level pressure, and daily minimum, mean, and maximum temperature.
We derive minicube-level NDVI from the B04 and B8A bands after applying the provided quality masks, yielding a sparse sequence of valid clear-sky vegetation observations, while meteorological variables remain available on a dense daily grid.
Weather covariates are augmented with cumulative rainfall and temperature-stress indicators computed over short temporal windows and between consecutive satellite observations.
Static context comprises geographic coordinates and the K\"oppen--Geiger climate class.
We follow the official GreenEarthNet validation and spatial, temporal, and joint spatio-temporal out-of-distribution (OOD) splits.
The future meteorological sequence defines the forcing used by VegSim and can be replaced at inference time to generate scenario-conditioned rollouts.

\paragraph{\textbf{Problem formulation.}}
For each minicube, we observe a sparse history
$\{(y_t,\mathbf{x}^{\mathrm{hist}}_t,\tau_t)\}_{t=1}^{T_h}$,
where $y_t \in [-1,1]$ is the NDVI value at acquisition time $\tau_t$ and $\mathbf{x}^{\mathrm{hist}}_t$ contains the associated
meteorological covariates.
The model also receives static spatial metadata
$\mathbf{s}=(\mathbf{s}^{\mathrm{geo}},s^{\mathrm{clim}})$ and a dense
future forcing sequence
$\{\mathbf{x}^{\mathrm{fut}}_k\}_{k=1}^{L}$.
Each rollout step $k$ is associated with a lead time $\Delta t_k$,
while future NDVI observations are available only at a sparse subset
of steps.
VegSim predicts conditional NDVI quantiles
\begin{equation}
q_{k,a} \approx
Q_a\!\left[
y_k
\mid
y_{1:T_h},
\mathbf{X}^{\mathrm{hist}},
\mathbf{X}^{\mathrm{fut}},
\Delta t_{1:L},
\mathbf{s}
\right],
\qquad
a \in \{0.1, 0.5, 0.9\}.
\label{eq:problem}
\end{equation}

\paragraph{\textbf{Model architecture.}}
\autoref{fig:architecture} summarizes the VegSim architecture.
The history encoder $E_{\mathrm{hist}}$ projects the sparse sequence of historical NDVI observations and paired meteorological covariates to the model dimension and processes it with a Transformer encoder using sinusoidal positional encodings.
Learned-query attention pooling compresses the valid historical tokens into the initial latent vegetation state $\mathbf{z}_0$.
The future encoder $E_{\mathrm{fut}}$ combines the dense future meteorological forcing with a learned lead-time embedding $\mathbf{e}_{\Delta t_k}$ and processes the resulting sequence with a second Transformer encoder, yielding future tokens $\mathbf{f}_k$.
In parallel, the spatial encoder $E_{\mathrm{spatial}}$ represents geographic coordinates $\mathbf{s}^{\mathrm{geo}}$ through harmonic features and the K\"oppen--Geiger climate class $s^{\mathrm{clim}}$ through a learned embedding, combining them into a static spatial context vector $\mathbf{c}_s$.

At each rollout step, the future token, spatial context, and lead-time embedding are combined to update the latent vegetation state:
\begin{equation}
\begin{aligned}
\mathbf{u}_k &=
\mathrm{MLP}_{\mathrm{in}}
([\mathbf{f}_k \Vert \mathbf{c}_s \Vert \mathbf{e}_{\Delta t_k}]),\\
\mathbf{z}_k &=
\mathrm{GRUCell}(\mathbf{u}_k,\mathbf{z}_{k-1}),
\qquad
\hat{\mathbf{q}}_k =
\mathrm{Dec}([\mathbf{z}_k \Vert \mathbf{c}_s]).
\end{aligned}
\label{eq:vegsim_rollout}
\end{equation}

The decoder is shared across rollout steps and predicts the $0.1$, $0.5$, and $0.9$ NDVI quantiles.
The latent state is propagated over the full future horizon, while supervision is applied only at steps with valid future satellite observations.
The model is trained with a temporally weighted pinball loss, where targets at lead time $\Delta t_k$ are weighted as
$w_k=(1+\alpha\Delta t_k)^{-1}$.
The total objective is
\begin{equation}
\mathcal{L}=\mathcal{L}_{\mathrm{pin}}+\lambda_{\mathrm{nc}}\mathcal{L}_{\mathrm{nc}},
\end{equation}
where $\mathcal{L}_{\mathrm{nc}}$ is a soft penalty that discourages quantile crossing.

\paragraph{\textbf{Scenario simulation.}}
At inference time, a meteorological scenario is defined by a set of future covariate channels, a temporal window, a perturbation rule, and the corresponding magnitudes.
Perturbations are specified in physical units before normalization, using additive changes for temperature variables and multiplicative changes for precipitation.
For a given scenario, the historical observations and static spatial context are kept fixed, so that the base and perturbed rollouts share the same inferred initial state $\mathbf{z}_0$ and differ only in the future meteorological forcing.
When engineered meteorological features are used, they are recomputed from the perturbed raw sequence before normalization.
The perturbed forcing is then propagated through the same trained latent dynamics without retraining, yielding scenario quantiles $\hat{q}^{\mathrm{scen}}_{k,a}$.
We quantify the scenario response relative to the unperturbed forcing as
\begin{equation}
\Delta q_{k,a}
=
\hat{q}^{\mathrm{scen}}_{k,a}
-
\hat{q}^{\mathrm{base}}_{k,a},
\qquad a \in \{0.1,0.5,0.9\},
\label{eq:scenario_response}
\end{equation}
where $\hat{q}^{\mathrm{base}}_{k,a}$ denotes the corresponding quantile under the unperturbed forcing.

\begin{table*}[t]
\caption{
Quantitative comparison across GreenEarthNet evaluation splits (val, ood-s, ood-t, and ood-st).
Lower values are better for both MAE and CRPS.
\colorbox{lightgreen}{\textbf{Bold}} values indicate the best result for each split and metric.
}
\centering
\setlength{\tabcolsep}{4.5pt}
\renewcommand{\arraystretch}{1.15}

\resizebox{\textwidth}{!}{%
\begin{tabular}{lcccccccc|cc}
\toprule

& \multicolumn{4}{c}{\textbf{MAE} $\downarrow$}
& \multicolumn{4}{c}{\textbf{CRPS} $\downarrow$}
& \textbf{Params}
& \textbf{MFLOPs} \\

\cmidrule(lr){2-5}
\cmidrule(lr){6-9}

\textbf{Model}
& \textbf{val}
& \textbf{ood-s}
& \textbf{ood-t}
& \textbf{ood-st}
& \textbf{val}
& \textbf{ood-s}
& \textbf{ood-t}
& \textbf{ood-st}
& \textbf{(M)}
& \\

\midrule

LSTM~\cite{hochreiter1997long}
& $.088_{\pm .085}$
& $.097_{\pm .101}$
& $.077_{\pm .081}$
& $.089_{\pm .091}$
& $.055_{\pm .060}$
& $.062_{\pm .072}$
& $.049_{\pm .057}$
& $.057_{\pm .065}$
& 3.99
& 208.20 \\

TimeLLM~\cite{jin2023time}
& $.075_{\pm .085}$
& $.101_{\pm .112}$
& $.072_{\pm .084}$
& $.086_{\pm .094}$
& --
& --
& --
& --
& 124.44
& 510.61 \\

InceptionTime~\cite{ismail2020inceptiontime}
& $.092_{\pm .081}$
& $.103_{\pm .100}$
& $.082_{\pm .078}$
& $.089_{\pm .085}$
& $.068_{\pm .064}$
& $.078_{\pm .079}$
& $.060_{\pm .061}$
& $.067_{\pm .068}$
& 1.23
& 64.42 \\

iTransformer~\cite{liu2024itransformer}
& $.084_{\pm .080}$
& $.106_{\pm .106}$
& $.078_{\pm .079}$
& $.091_{\pm .090}$
& --
& --
& --
& --
& 4.22
& 8.43 \\

TimeXer~\cite{wang2024timexer}
& $.075_{\pm .095}$
& $.096_{\pm .120}$
& $.066_{\pm .087}$
& $.081_{\pm .100}$
& --
& --
& --
& --
& 6.33
& 21.06 \\

Chronos-2~\cite{ansari2025chronos2}
& $.086_{\pm .111}$
& $.110_{\pm .138}$
& $.077_{\pm .105}$
& $.092_{\pm .119}$
& $.057_{\pm .069}$
& $.074_{\pm .091}$
& $.052_{\pm .068}$
& $.062_{\pm .079}$
& 28
& 316.87 \\

Contextformer~\cite{benson2024multimodal}
& $.121_{\pm .031}$
& $.129_{\pm .038}$
& $.119_{\pm .041}$
& $.123_{\pm .039}$
& --
& --
& --
& --
& 6.1
& $173 \times 10^{3}$ \\

\midrule

VegSim
& \cellcolor{lightgreen}$\mathbf{.057_{\pm .051}}$
& \cellcolor{lightgreen}$\mathbf{.077_{\pm .066}}$
& \cellcolor{lightgreen}$\mathbf{.060_{\pm .057}}$
& \cellcolor{lightgreen}$\mathbf{.077_{\pm .065}}$
& \cellcolor{lightgreen}$\mathbf{.035_{\pm .034}}$
& \cellcolor{lightgreen}$\mathbf{.047_{\pm .043}}$
& \cellcolor{lightgreen}$\mathbf{.037_{\pm .038}}$
& \cellcolor{lightgreen}$\mathbf{.047_{\pm .043}}$
& 1.3
& 3.28 \\

\bottomrule
\end{tabular}
}
\label{tab:metrics}
\end{table*}

\section{Experimental Setup}
\label{sec:experimental}

\paragraph{\textbf{Training Details.}}
The history and future encoders use four Transformer layers with
$d_{\mathrm{model}}=128$; the latent state has dimension $d_z=128$,
while the lead-time and spatial embeddings have dimensions 32.
VegSim is trained with Adam~\cite{kingma2014adam} for 200 epochs
with batch size 128 and an initial learning rate of $10^{-4}$.
The temporal weighting coefficient is set to $\alpha=0.1$, and the non-crossing penalty weight to $\lambda_{\mathrm{nc}}=10^{-2}$.
All experiments run on a single NVIDIA T4 GPU.

\paragraph{\textbf{Baselines and Evaluation Protocol.}}
Scenario-conditioned simulations cannot be evaluated directly against ground truth, since vegetation observations under perturbed meteorological forcing are unavailable.
We therefore assess the learned dynamics through forecasting under observed future forcing before applying them to perturbed forcing sequences.
We compare VegSim with LSTM~\cite{hochreiter1997long}, InceptionTime~\cite{ismail2020inceptiontime}, iTransformer~\cite{liu2024itransformer}, TimeXer~\cite{wang2024timexer}, TimeLLM~\cite{jin2023time}, Chronos-2~\cite{ansari2025chronos2}, and the EO-specific Contextformer~\cite{benson2024multimodal}.
All models are evaluated on the GreenEarthNet val, ood-s, ood-t, and ood-st splits.
We report MAE for point accuracy and CRPS for models providing probabilistic predictions; both metrics are minimized.

\begin{figure}
    \centering
    \includegraphics[width=\linewidth]{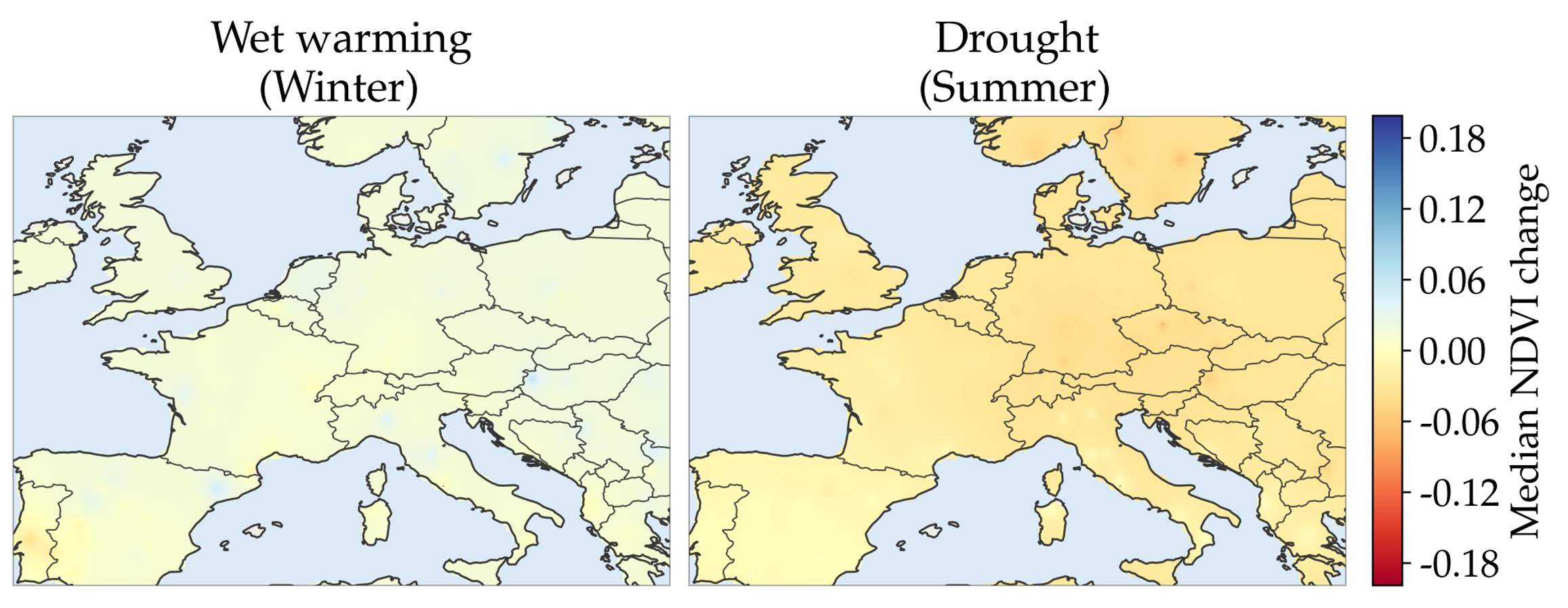}
\caption{Spatial distribution of the median NDVI response ($\Delta$NDVI) under Wet warming (winter, $\Delta T=+1.5\,^\circ\mathrm{C}$, $\Delta P=+15\%$) and Drought (summer, $\Delta T=+2\,^\circ\mathrm{C}$, $\Delta P=-40\%$). Responses are computed relative to the unperturbed forcing and aggregated across all GreenEarthNet evaluation subsets.}
\Description{Two maps of Europe showing the median NDVI response simulated by VegSim under two meteorological perturbation scenarios: wet warming in winter and drought in summer. The color scale represents the median NDVI change relative to the unperturbed forcing, with negative values indicating lower simulated NDVI and positive values indicating higher simulated NDVI.}
    \label{fig:scenarios}
\end{figure}

\section{Results}
\label{sec:results}

\paragraph{\textbf{Forecasting performance.}}
\autoref{tab:metrics} compares VegSim with the forecasting baselines across the GreenEarthNet evaluation splits.
VegSim achieves the lowest MAE in all four regimes, with errors remaining close to validation performance under temporal shift and increasing mainly under spatial and joint spatio-temporal shifts.
Among models providing probabilistic outputs, VegSim also obtains the lowest CRPS across all splits.
Relative to the strongest competing baseline in each split, VegSim reduces MAE by $24\%$, $20\%$, $9\%$, and $5\%$ on val, ood-s, ood-t, and ood-st, respectively; the corresponding CRPS reductions relative to the best probabilistic baseline are $36\%$, $24\%$, $24\%$, and $18\%$.
One-sided paired Wilcoxon signed-rank tests on source-level metric differences, with Holm--Bonferroni correction within each split, support the main performance gains.
VegSim achieves this performance with 1.3M parameters and 3.28 MFLOPs, lower model complexity and computational cost than most competing models.

\paragraph{\textbf{Scenario-conditioned simulation.}}
We evaluate VegSim under two contrasting meteorological scenarios applied to their target seasons: Wet warming (winter, $\Delta T=+1.5\,^\circ\mathrm{C}$, $\Delta P=+15\%$) and Drought (summer, $\Delta T=+2\,^\circ\mathrm{C}$, $\Delta P=-40\%$).
\autoref{fig:scenarios} reports the spatial distribution of the median NDVI response relative to the unperturbed forcing, aggregated across all evaluation subsets.

\textit{Wet warming} (winter, $\Delta T = +1.5\,^\circ\mathrm{C}$, $\Delta P = +15\%$).
Winter warming produces a weak but broadly positive response across the study area.
Low temperatures constrain vegetation activity during the cold season, and the warmer and wetter forcing shifts the simulated vegetation trajectory toward higher NDVI.
The response remains relatively small in magnitude and spatially smooth across most of Europe.

\textit{Drought} (summer, $\Delta T = +2\,^\circ\mathrm{C}$, $\Delta P = -40\%$).
A strong summer precipitation deficit combined with warmer temperatures produces a widespread negative response across Europe.
The simulated NDVI decreases across most of the study area, with limited spatial differentiation compared with the overall magnitude of the response.
At this level of meteorological stress, the negative response extends across different climatic regions rather than being confined to a localized area.

The opposite responses show that VegSim propagates the direction and seasonal context of the imposed meteorological forcing through its latent vegetation dynamics.
These outputs are conditional simulations under perturbed forcing, not causal estimates of meteorological effects.

\section{Conclusion}
\label{sec:conclusion}

We introduced VegSim, a geospatial world model that combines sparse satellite-derived vegetation observations, dense meteorological forcing, and static spatial context to support both probabilistic forecasting and scenario-conditioned simulation.
Across the GreenEarthNet evaluation splits, VegSim achieves strong point and probabilistic forecasting accuracy while retaining a compact computational footprint.
Under controlled meteorological perturbations, the same learned dynamics produce distinct vegetation responses to winter wet warming and summer drought, showing that future weather can be used as a controllable forcing input without retraining the model.
These rollouts are conditional simulations rather than causal estimates, and their reliability depends on the learned dynamics and the support of the training distribution.
Future work will investigate broader meteorological scenarios, uncertainty under perturbed forcing, and interactive tools for user-defined scenario analysis.

\bb
\begin{acks}
Irene Iele is a Ph.D. student enrolled in the National Ph.D. in Artificial Intelligence, course on Health and Life Sciences, organized by Università Campus Bio-Medico di Roma.
We acknowledge the EuroHPC Joint Undertaking for granting this project access to the EuroHPC supercomputer Vega, hosted by the Institute of Information Science (Slovenia), under a EuroHPC Development Access call.
\end{acks}

\bibliographystyle{ACM-Reference-Format}
\bibliography{references}
\end{document}